\documentclass{article}

\usepackage{arxiv}

\usepackage[utf8]{inputenc} % allow utf-8 input
\usepackage[T1]{fontenc}    % use 8-bit T1 fonts
\usepackage{hyperref}       % hyperlinks
\usepackage{url}            % simple URL typesetting
\usepackage{booktabs}       % professional-quality tables
\usepackage{amsfonts}       % blackboard math symbols
\usepackage{nicefrac}       % compact symbols for 1/2, etc.
\usepackage{microtype}      % microtypography
\usepackage{graphicx} % Package for handling images
\graphicspath{{./images/}} % Set path for image folder
\usepackage{float} % Package for controlling floating objects like figures and tables
\usepackage[backend=bibtex]{biblatex} % Package for bibliography
\title{Deriving a Quantitative Relationship Between Resolution and Human Classification Error}

\author{
  Josiah I.Clark\\
  HSB Intel\\
  Reston, VA 20190 \\
  \texttt{josiah@hsbintel.com} \\
   \And
 Caroline A. Clark \\
  HSB Intel\\
  Reston, VA 20190 \\
  \texttt{caroline@hsbintel.com} \\
}

\begin{document}
\maketitle

\begin{abstract}
For machine learning perception problems, human-level classification performance is used as an estimate of top algorithm performance.  Thus, it is important to understand as precisely as possible the factors that impact human-level performance.  Knowing this 1) provides a benchmark for model performance, 2) tells a project manager what type of data to obtain for human labelers in order to get accurate labels, and 3) enables ground-truth analysis--largely conducted by humans--to be carried out smoothly.  In this empirical study, we explored the relationship between resolution and human classification performance using the MNIST data set down-sampled to various resolutions.  The quantitative heuristic we derived could prove useful for predicting machine model performance, predicting data storage requirements, and saving valuable resources in the deployment of machine learning projects.  It also has the potential to be used in a wide variety of fields such as remote sensing, medical imaging, scientific imaging, and astronomy.
\end{abstract}

% keywords
\keywords{object resolution \and optimal object resolution \and machine learning \and feature extraction \and computer vision \and machine vision \and deep learning \and artificial intelligence \and data labeling \and NIIRS \and resolution \and image quality \and remote sensing \and ground truth \and human-level performance \and human labeling performance}

\section{Introduction}
Although there have been many exciting breakthroughs in machine learning, shipping new AI products is still hard.  A top challenge organizations face is acquiring sufficient labeled data with the attributes necessary to achieve their target trained model performance objectives \cite{hansen_mapping_2019, mnih_learning_2012, wang_deep_2019, zhu_pick_2019}.  For visual perception problems, a key attribute that organizations must choose is the resolution requirement of the training data.

The resolution of training data directly influences and sets the parameters of a variety of deep learning related tasks:
    \begin{enumerate}
        \item In many fields (e.g. remote sensing, medical imaging, astronomy), the cost of data acquisition increases exponentially relative to image resolution.
        \item Storage capacity requirements are squared for every doubling in data resolution.
        \item Model width is usually approximately equal to the input image resolution.
        \item Gradient Descent computations increase as training data resolution increases.
        \item Trained model size increases as resolution increases.
        \item Inference time can increase as model size increases.
    \end{enumerate}

The effect of the above is that many organizations have many good reasons to err on the side of collecting lower resolution images and few good reasons to err on the side of higher resolution.  Unfortunately, this can have dire impacts on the maximum performance of deep learning models trained on these images.

For visual perception problems, human classification performance comes close to approximating the theoretical maximum trained model performance, or Bayes error \cite{ng_deeplearningai}. Thus, it could be said that, for visual perception problems:
% \vspace{-1em}
    \begin{center}
        $ \text{human error} \approx \text{Bayes error} \approx \text{maximum trained model performance}$
    \end{center}

Understanding the quantitative relationship between human classification error and image/object resolution for a given data set would enable a deep learning practitioner to determine the optimal training data resolution given a target model performance. Unfortunately, after extensive review of the literature, the most relevant methods for estimating human classification performance given a specific image were the NIIRS method and the resolution method \cite{irvine_quant_2007, irvine_tailoring_2018, riehl_comparison_1996, walcz_eval_2019}.

If such a quantitative relationship between human classification error and resolution were derived, it would enable a user to:
    \begin{enumerate}
        \item Estimate model error when an existing data source is being considered for usage as the training set.
        \item Estimate the object/image resolution required to achieve a target model performance goal.
    \end{enumerate}

In this paper, we review the current literature related to predicting human classification error given an image.  We discuss the need for a method applicable to deep learning for predicting human classification performance, and how this would improve machine learning workflows.  We discuss the method we used to test human classification performance on randomized, down-sampled MNIST images.  We review the results of our experiment, how they compare to existing methods, and recommendations for future study. 

%%% Literature Review
\section{Literature Review}
    %%% Machine Learning Model Error, Human Error, and Bayes Error
    \subsection{Machine Learning Model Error, Human Error, and Bayes Error}
     Given a data source of sufficient fidelity, humans are very good at perception problems.  In machine learning, perception problems generally can be described as training a machine to classify a system based on a human labeled training set, where the human leverages their senses to label the desired observations of a class.  Thus the goal of a solving a perception problem with machine learning is to build a deep neural network capable of perceiving the desired class in a similar fashion to the human labeler.

    \begin{figure}[H]
        \centering
        \includegraphics[scale=0.23]{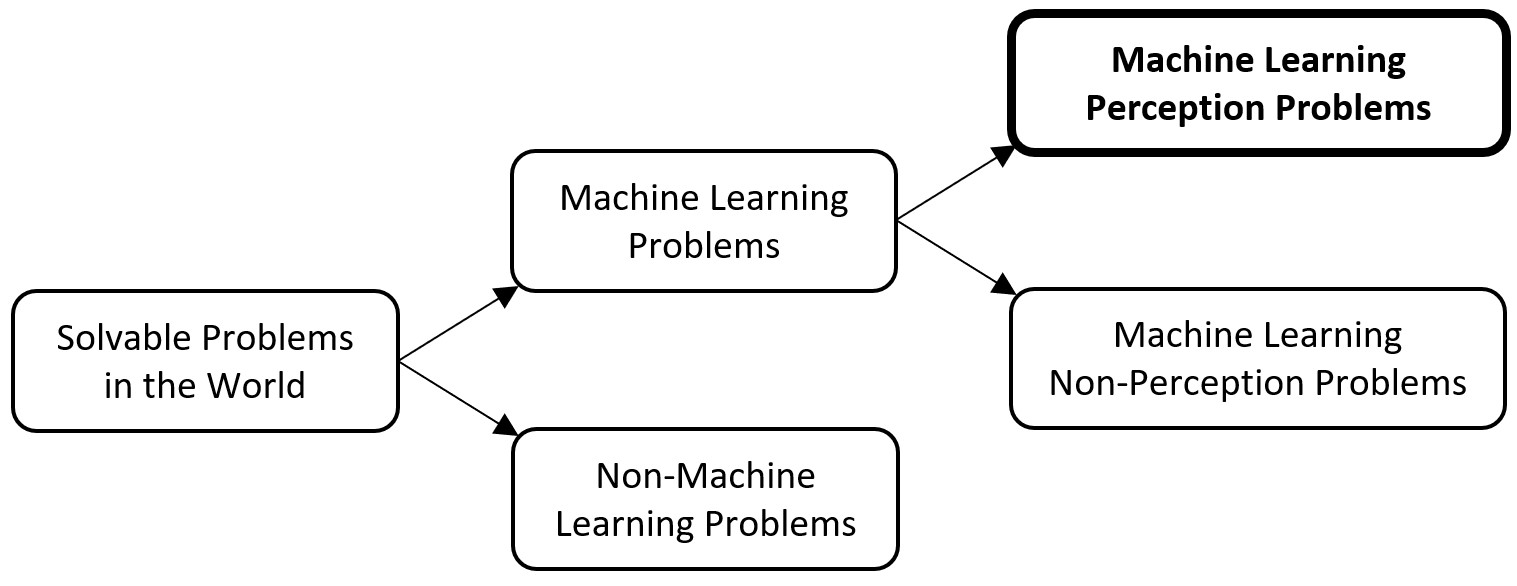}
        \caption{Perception problems as a subset of problems solvable with machine learning.}
        \label{fig:mlproblems}
    \end{figure}
    
   It is substantially easier to to train deep learning models up to the level of human performance on a given data source. This is because typically humans are labeling data and conducting ground-truth analysis.
    \vspace{-0.5em}
    \begin{figure}[H]
        \centering
        \includegraphics[scale=0.52]{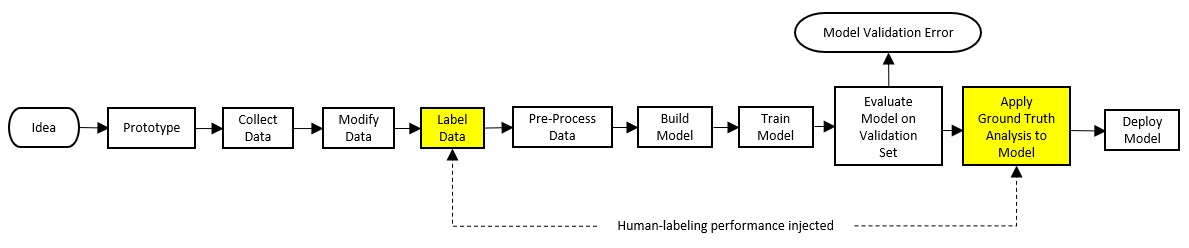}
        \caption{Injection points of human labeling performance in the machine learning workflow.}
        \label{fig:mlprocess3}
    \end{figure}
    
    In practice, the lowest achievable error rate by a human expert is considered a good approximation of Bayes error.  For example, most humans are experts at recognizing classes such as human faces, cats, and cars.  In the case of interpretation of an x-ray, the lowest human error rate could be the average error rate of the top five specialists in the field.  Once a model surpasses human-level performance, it becomes more difficult to improve performance, though not impossible. 
    
    \begin{figure}[H]
        \centering
        \includegraphics[scale=0.35]{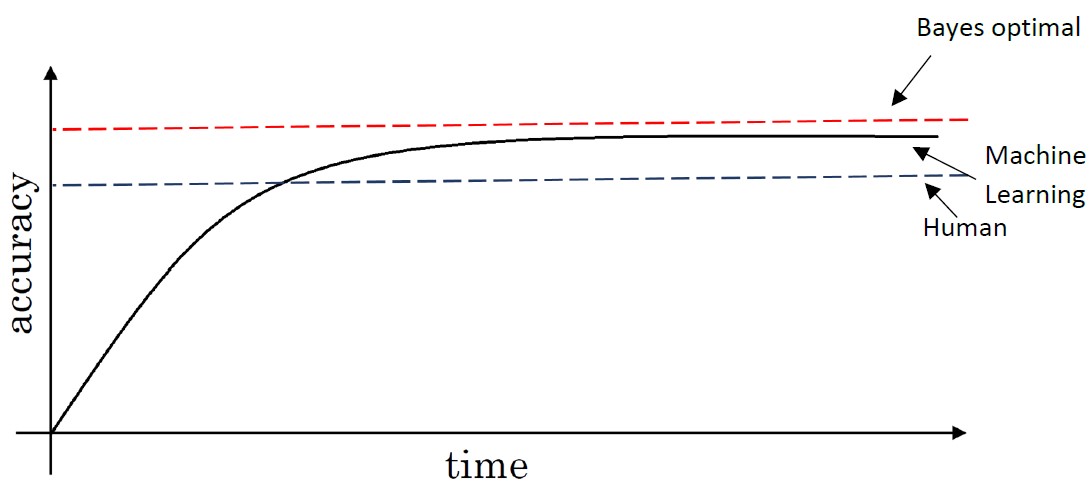}
        \caption{Machine learning performance, human performance, and Bayes error for perception problems \cite{ng_deeplearningai}.}
        \label{fig:human_perf}
    \end{figure}

    %%% Evaluating Image Quality
    \subsection{Evaluating Image Quality}
    Much research has been done on how to rate image quality, and thus its interpretability and usefulness, over the years.  Riehl \& Maver (1996) \cite{riehl_comparison_1996} compared the long-used resolution method and the National Imagery Interpretability Rating Scale (NIIRS) \cite{civil_niirs} as common aerial reconnaissance image quality measures. Resolution is defined in this paper as "indicative of the smallest object that can be detected", and "relates to separating, making visible, or distinguishing small detail" \cite[p.242]{riehl_comparison_1996}.
    
    Resolution in this sense is measurable in the real world and a laboratory with a tri-bar target test. It is possible to say that an image is lower or higher resolution, and rate its interpretability. However, it stops short of quantifying a relationship between resolution and image quality or interpretability.
    
    The NIIRS method, while robust to the technical shortcomings that caused the resolution method to fall out of favor, also does not quantify a relationship between resolution and image quality or interpretability. Rather, it assigns a NIIRS score to a given image based on interpretability by a highly trained specialist. 
    
    The process used to develop NIIRS is complex, resource-intensive, and requires rigorous methods \cite{irvine_national_1997}.  However, NIIRS is still considered state-of-the-art in aerial image quality and interpretability \cite{gerwe_application_2012, irvine_quant_2007, irvine_tailoring_2018, walcz_eval_2019}, and is being extended to other media \cite{blasch_niirs_2015}.

    %%% Resolution in Machine Vision and Required Pixels
    \subsection{Resolution in Machine Vision and Required Pixels}\label{CV}
    In contrast, the field of machine and computer vision presents a quantitative approach to resolution.  In \textit{Handbook of Machine and Computer Vision: A Guide for Developers and Users} \cite{hornberg_2017}, the authors recommend the following method to determine the required camera resolution ($Rc$) for a given machine vision task:
        \begin{figure}[!ht]
            \centering
            \includegraphics[scale=0.45]{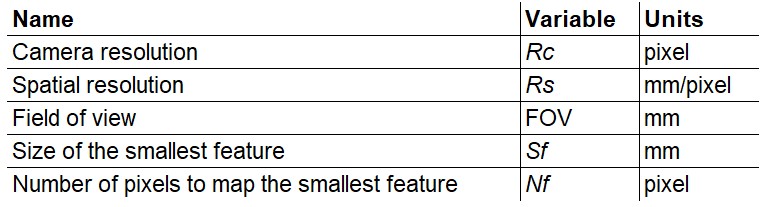}
            \label{fig:res_calc}
        \end{figure}
    \begin{equation}
        Rc = \frac{FOV}{Rs} = FOV \times \frac{Nf}{Sf}
    \end{equation}
    The issue here is that a gap exists in selecting or determining the number of pixels needed to map the smallest feature, in order to achieve the given task. 
    
    Cai (2003) \cite{cai_how_2003} investigated the topic of required pixels for image recognition by associating different levels of pixel requirements with different image content categories such as faces and outdoor scenes. However, this experiment did not take into account the differing scales of the images, and it is therefore difficult to interpret and generalize the findings.

    %%% The Need for a Deep Learning-Specific Method of Evaluating Image Quality
    \subsection{The Need for a Deep Learning-Specific Method of Evaluating Image Quality}
    There is broad application in the commercial space for a method that would enable machine learning teams to rapidly quantify potential model feature extraction capabilities on an available data set, or determine data set requirements given a feature extraction performance objective.
    
\begin{figure}[H]
        \centering
        \includegraphics[scale=0.46]{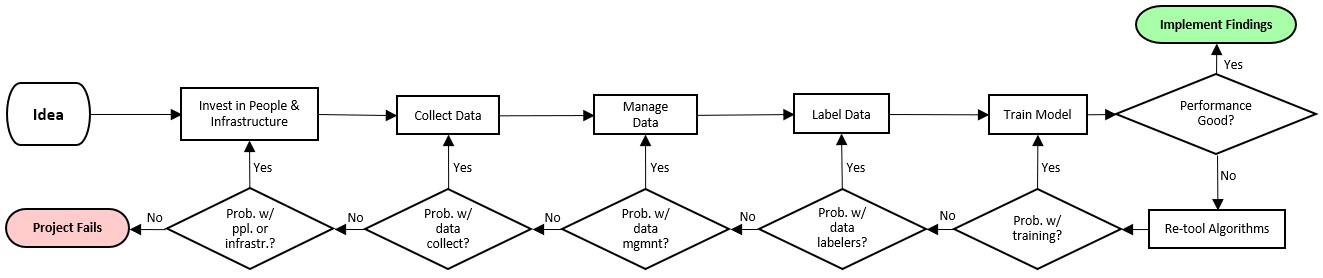}
        \caption{Current state of the art machine learning project process.}
        \label{fig:mlprocess1}
    \end{figure}

    Machine learning and AI is positioned to transform every industry\cite{ng_transformation_2018}. Private and public sector leaders are looking into how they can leverage this growing field of research to improve their offerings and capabilities. It is commonly accepted that leaders must invest in acquiring, managing, and labeling data in order to benefit from ML/AI, and that they must avoid acquiring low-value data and wasting resources. Even so, the current state of the art method for executing a machine learning project is show in Figure \ref{fig:mlprocess1}.
    
    Private or public entities might start with an idea for using deep learning; obtain data they believe suits the purpose; collect, manage, and label the data; and train models on the data.  It is only after this last step that performance is evaluated, and improvements are attempted.
    
\begin{figure}[H]
        \centering
        \includegraphics[scale=0.46]{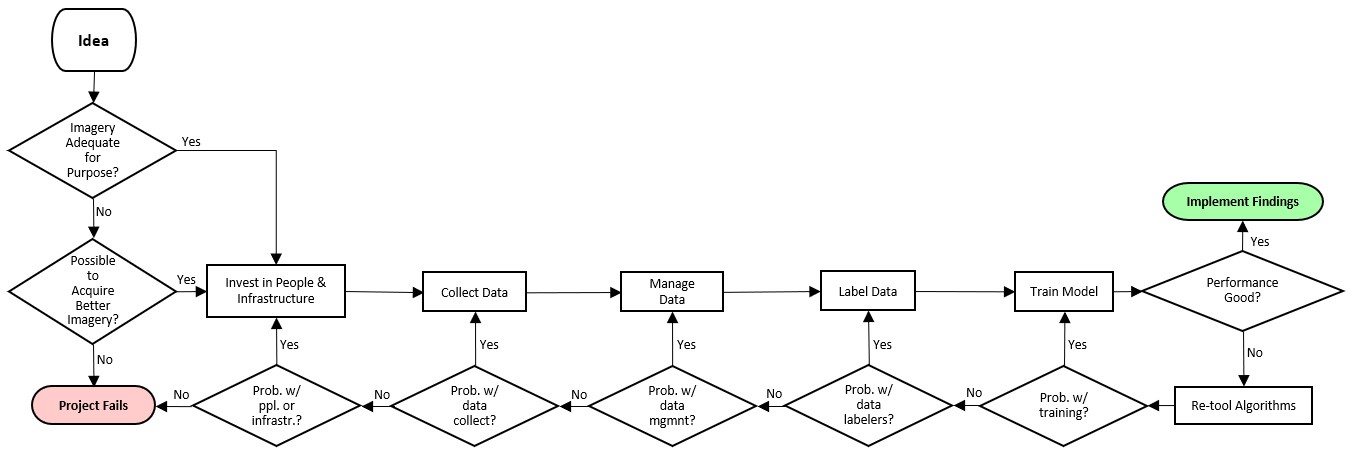}
        \caption{Proposed machine learning project process.}
        \label{fig:mlprocess2}
    \end{figure}

    Although this workflow is common, the process of learning through trial and error can be costly, slow projects down, and in some cases ultimately lead to failure.  The challenge that exists today is that there are few heuristics that can predict machine learning model performance prior to training.  If such a heuristic exists, organizations could calculate their data requirements or predict model performance prior to conducting trial and error analysis.  Additionally, armed with the knowledge of predicted model performance and data adequacy, organizations could focus their troubleshooting efforts on more likely culprits thus reducing solution delivery time and cost.  Figure \ref{fig:mlprocess2} depicts the process updated if such a heuristic were found.
    
    Interested parties could in theory utilize an optimal object resolution process alone to determine if they have the right data set to meet their feature extraction objectives.  This would remove the current requirement for machine learning teams to conduct expensive imagery collection and or labeling prior to determining the performance potentials of their trained models.  Precluding this initial step would provide substantial cost savings especially in situations where the required imagery to meet the target objective does not exist.
        
    In order to be useful, the image data evaluation step needs to:
        \begin{enumerate}
            \item Be low cost
            \item Be applicable to wide range of problems
            \item Be easy to use
            \item Require as little training should be required
            \item Provide good estimates even on small data sets
        \end{enumerate}

    As discussed above, both the resolution and NIIRS methods do not provide a relationship for quantifying the required number of pixels per object to achieve a given image interpretation objective.  They also both require specially trained imagery analysts to assess image quality, driving up costs. Finally, the application of these image quality methods does not extend beyond remote sensing to industries such as medical imaging, scientific imagine, or astronomy.
    
    \begin{figure}[H]
        \centering
        \includegraphics[scale=0.35]{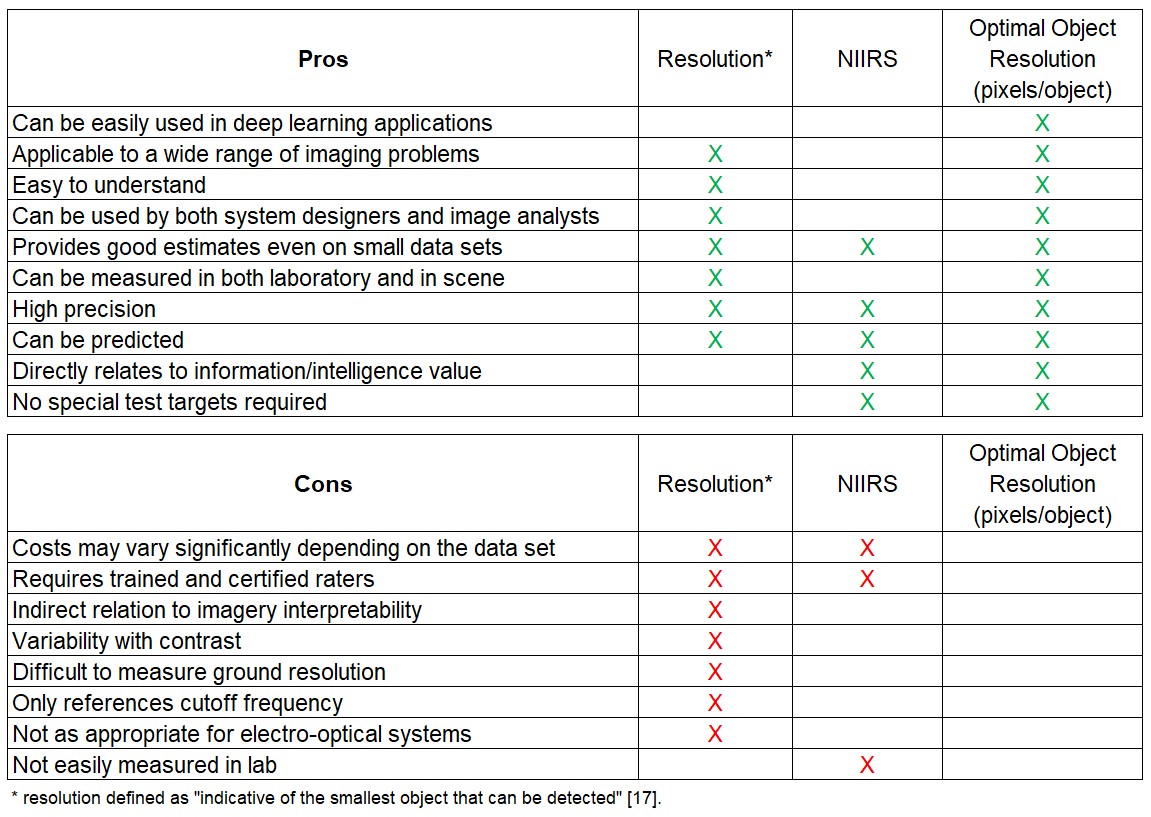}
        \caption{Comparison of methods for evaluating image quality.}
        \label{fig:table_comp}
    \end{figure}
    
    Because of the significant investment public and private entities make in acquiring, labeling, and processing data before getting feedback model performance metrics, it's crucial that there exists a method for predicting success of an ML project before data is acquired.  If a method for deriving the required resolution were developed, this could potentially be achieved. This could also move us toward a more universal definition of the variable \textit{Nf} (the number of pixels required to map the smallest feature) in the equation presented in section 2.3 when applied to human or machine vision problems respectively.

%%% Methodology
\section{Methodology}

    %%% Experiment Objective
    \subsection{Experiment Objective}
    Our objective was to profile human image classification performance as a function of object resolution. 
    
    %%% Experiment Requirements
    \subsection{Experiment Requirements}
        \begin{enumerate}
            \item Problem is a visual perception problem.
            \item Typical humans must be experts at the classification objective.
            \item The test must be randomized.
            \item The outcome must be numerically quantifiable.
            \item The test set must be publicly available.
            \item Each object must have the same relative scale within each example.
            \item The human test must be repeatable with a machine learning algorithm.
        \end{enumerate}
    
    %%% Experiment Approach
    \subsection{Experiment Approach}
    In order to achieve our objective and meet our requirements, we developed a Human Image Classification Error Analysis Application (HICEAA) using Python. We used the MNIST data set, which has 10 classes (0-9). We down-sampled these images so that they would range from 1x1 to 28x28, with 28 possible resolutions in total.

    We built our labeling application so that it tracked and modeled experts individually \cite{guan_who_2018}.  We developed our application with a simple-to-use, well-marked interface \cite{galizia_advanced_2015}.  We did not focus on our testers being statistically relevant to the general population for these reasons:
        \begin{enumerate}
            \item Humans tend to be highly consistent in their segmentation of images \cite{martin_database_2001}.
            \item Most humans are experts at recognizing the numbers 0-9.  For more complex image content, it may make sense to implement a method to factor out bias \cite{misra_seeing_2016}.
            \item In practical terms, it is more important for labelers of machine learning data to be experts and produce accurate data than to be representative of a general population.
        \end{enumerate}

    HICEAA initiated the following loop until the test subject closed the application window: 
    
    \begin{enumerate}
        \item Randomly selected image from MNIST Validation data set.
        \item Randomly resized selected image to a resolution value ranging between 1x1 - 28x28 pixels.\footnote{We used the cv2.resize using their INTER\_AREA function since we were planning on down-sampling the images from higher resolution images.  The primary reason we used the INTER\_AREA method vs other resizing functions was due to the INTER\_AREA method producing moire'-free results when conducting image decimation.  Also, though the INTER\_AREA function is only recommended for images of 4 bands or less, our image was gray-scale (single band).  Individuals considering using our method on hyperspectral or multi-spectral imagery may find better results with a different decimation function.}
        \item Counted the resized image pixels that are greater than zero (0 is the background pixel value).
        \item Initiated timer and displays the following to test subject:
            \begin{enumerate}
                \item "INSTRUCTIONS: Select the number you see, if you can't recognize the number select -1."
                \item A 3.25x3.25 inch projection of the resized image.
                \item A set of selection buttons ranging from -1 to 9 (11 selection options in total).
                \item Upon user selection the application records the results listed in Figure \ref{fig:table_resrec} to a results table.

                    \begin{figure}[H]
                        \centering
                        \includegraphics[scale=0.25]{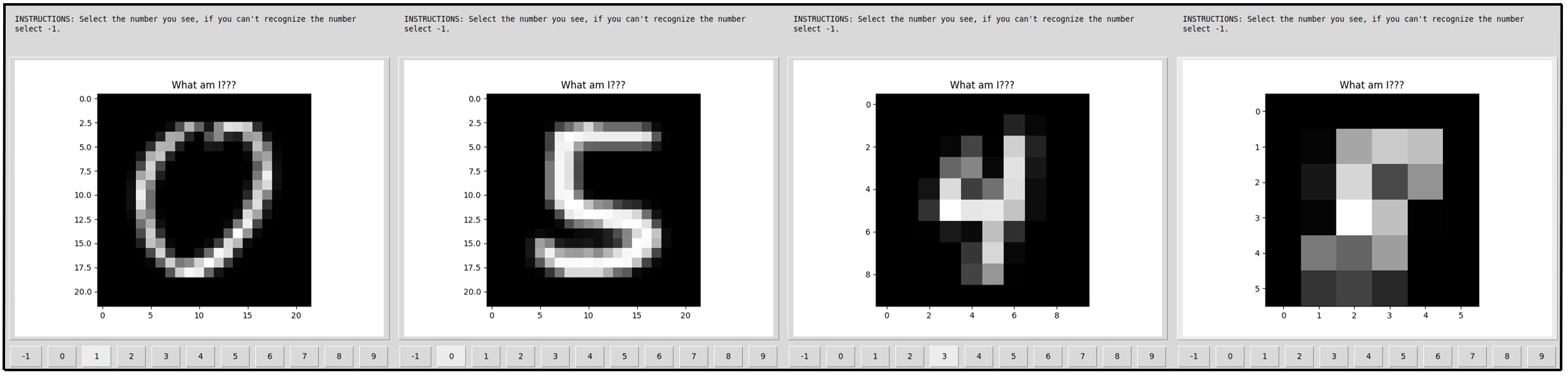}
                        \caption{Examples of down-sampled MNIST images presented to test subjects in the HICEAA.}
                        \label{fig:application}
                    \end{figure}
                    
                    \begin{figure}[H]
                        \centering
                        \includegraphics[scale=0.45]{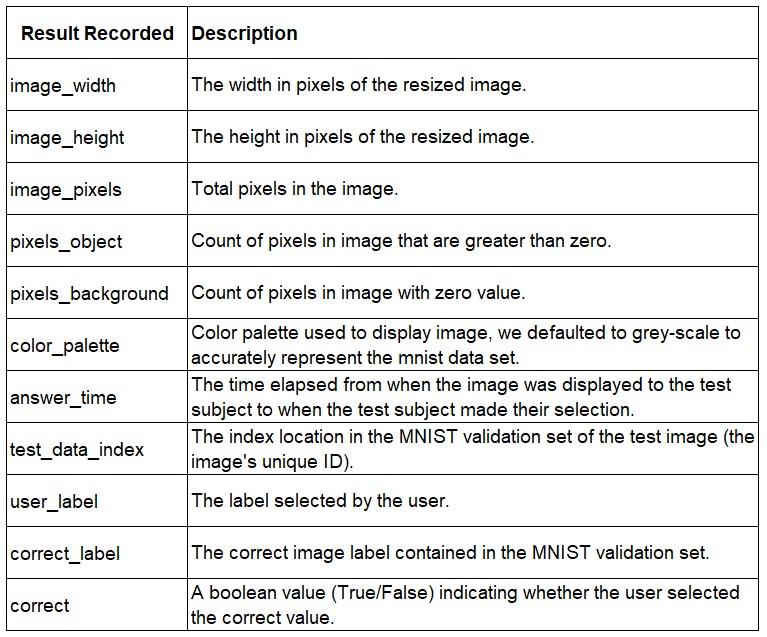}
                        \caption{Upon user selection the application records the following to a results table.}
                        \label{fig:table_resrec}
                    \end{figure}
            \end{enumerate}
    \end{enumerate}

    Upon exiting the application, a table containing relevant values by image resolution, and a table containing relevant values by pixels/object, is calculated:
            \begin{figure}[H]
                \centering
                \includegraphics[scale=0.45]{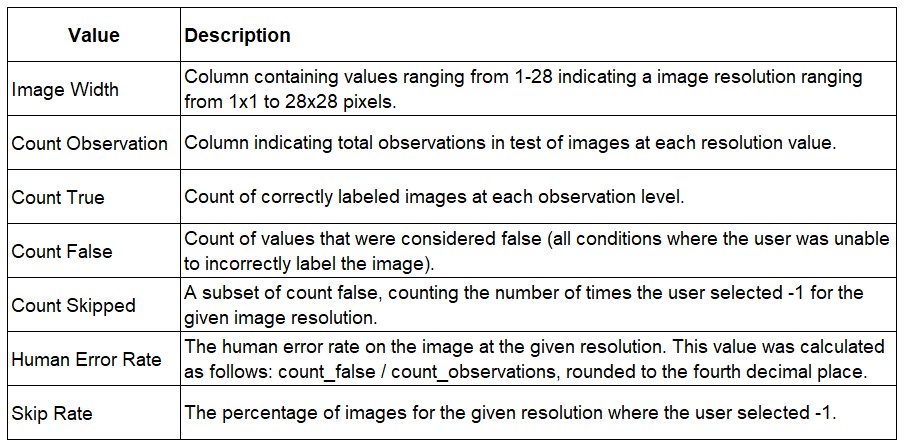}
                \caption{Aggregate error rate per image resolution table, calculated upon exiting the application.}
                \label{fig:table_imageres}
            \end{figure}            
            
            \begin{figure}[H]
                \centering
                \includegraphics[scale=0.45]{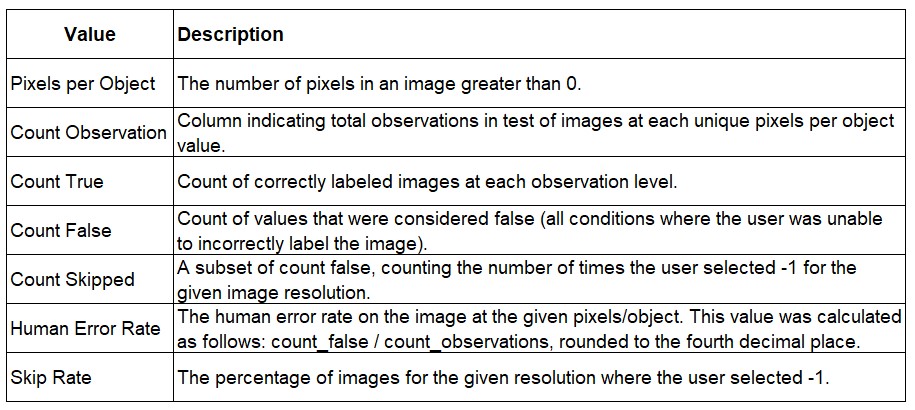}
                \caption{Aggregate error rate per number of pixels/object, calculated upon exiting the application.}
                \label{fig:table_pixobj}
            \end{figure}            

    10,000 labels from test subjects were collected using the HICEAA. The results were plotted and a regression analysis was conducted to analyze the quantitative relationship between image/object resolution and human classification error.

%%% Results
\section{Results}
We plotted the relationship between image resolution and human classification error rate.  We observed that an inverse sigmoid function described the relationship.
    \begin{figure}[H]
        \centering
        \includegraphics[scale=0.45]{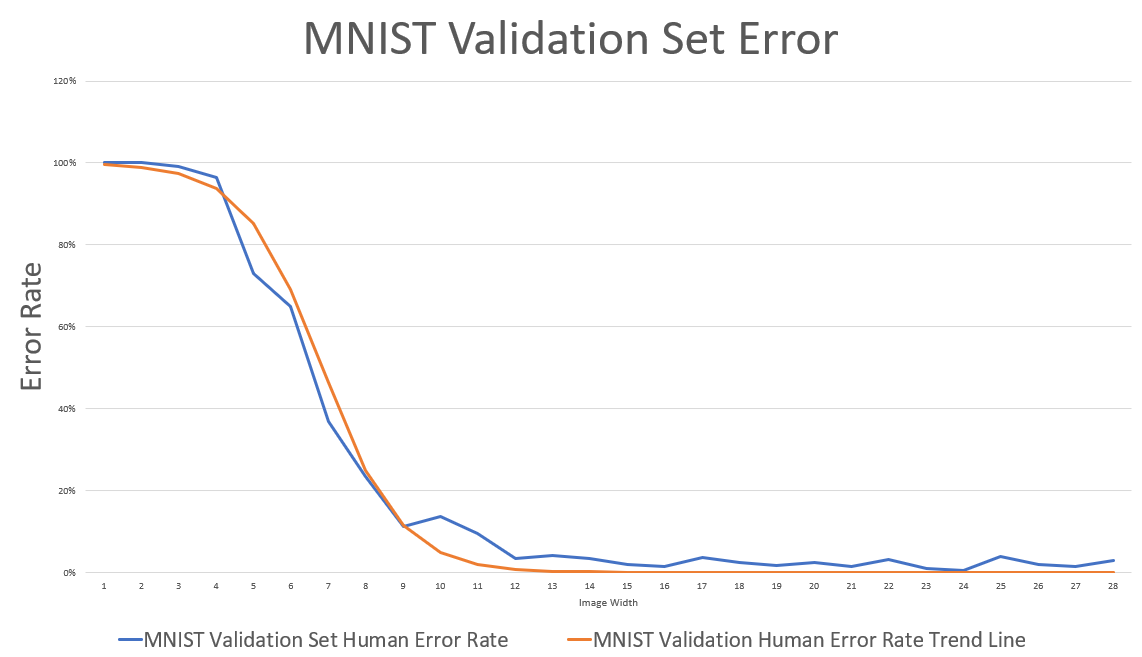}
        % \caption{}
        \label{fig:MNIST_validationerror}
    \end{figure}            

We arrived at the following relationship between resolution and human classification error, assuming that the image is equilateral (or square):\\
\newline
\newline
\forceindent \textit{Given:}\\
\forceindent$\text{sigmoid x-axis center} = c = 6.5$\\
\forceindent$\text{sigmoid accelerator} = \alpha = -0.95$\\
\forceindent$\text{image width} = x$\\
\begin{equation}
    \approx \text{human error rate} = y = f(x) = \frac{1}{1+e^{-(\alpha x + c)}}
\end{equation}

%%% Discussion
\section{Discussion}
Achieving human-level performance on perception tasks is a key goal of machine learning projects.  However, to-date a quantitative relationship between human classification performance and resolution has not yet been established, making it difficult to predict target model performance, as well as manage machine learning workflows.

In addition, current state-of-the-art methods for classifying images, resolution and NIIRS, do not meet all of the requirements of a solution to this problem for deep learning, most notably those related to cost, required training, and applicability to a variety of problems in different fields. 

In this paper we set out to profile human image classification performance as a function of object resolution.  We observed a quantifiable relationship between object resolution and human classification performance.  This heuristic could be leveraged to improve machine learning project workflows by using it to calculate data requirements or predict model performance prior to investment.  It could also help save valuable time troubleshooting model performance.  

Finally, this optimal object resolution, a value with the units pixels/object, could also be substituted for the variable $Nf$ in the computer vision formula in section \ref{CV}, defined as the "number of pixels to map the smallest feature".  We plan on doing further testing of this concept on additional data sets.

%%% Conclusion
\section{Conclusion}
In conclusion, we plan on utilizing our mathematical heuristic for "back of the napkin" machine learning project planning and utilizing the HICEAA for more precise human error predictions. This process should greatly reduce wasted time and effort and enable more accurate costing and feasibility analysis of visual perception machine learning projects prior to substantial investment.

\printbibliography

\end{document}